# L0 Regularization Based Neural Network Design and Compression

S. Asim Ahmed
Mission Systems
*Collins Aerospace*
Cedar Rapids, USA
asim.ahmed@collins.com

*Abstract—* We consider complexity of Deep Neural Networks (DNNs) and their associated massive over-parameterization. Such over-parametrization may entail susceptibility to adversarial attacks, loss of interpretability and adverse Size, Weight and Power - Cost (SWaP-C) considerations. We ask if there are methodical ways (regularization) to reduce complexity and how can we interpret trade-off between desired metric and complexity of DNN. Reducing complexity is directly applicable to scaling of AI applications to real world problems (especially for off-the-cloud applications). We show that presence of the knee of the tradeoff curve. We apply a form of L0 regularization to classifications of MNIST hand written digits and signal modulation data. We show that such regularization captures saliency in the input space as well.

## I. Introduction

Recent spectacular advances in machine learning has attracted attention towards size, weight and power considerations of these algorithms. For power consumption, we have an ultimate bench-mark of human brain that only consumes 20W. Managing hardware complexity is critical to deploying Machine Learning (ML) algorithms out to the edges. Theoretical ML framework like Probably Approximately Correct (PAC) learning shed light of sample complexity and pay no attention to computational complexity. In this paper, we will pursue an engineering centric approach of meeting design goals with resource constraints. We will ask if there is a methodical approach to trading-off complexity with desired metrics.

## II. Norm, Regulariztion and L0 Regularization

Fig. *1* shows mathematical balls for various p-norms, $||x|||_p$, in two dimensions. Notice that a unit ball with p-norms and p less than 1 results in a non-convex set. Also, set represented by $L_1$ norms is also non-smooth and non-differentiable everywhere.

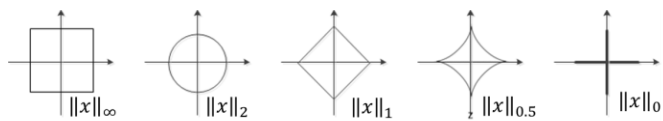

Fig. 1 Unit ball for various p-norms

Fig. 2 shows multi-objective optimization view of regularization with $L_1$ and $L_2$ norms. Here the intersection of Lp ball with loss function (red contours) determines the solution and we see why $L_1$ induces a sparser solution.

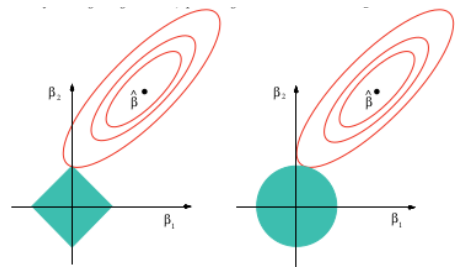

Fig. 2 Multi-Objective Optimization

The empirical risk with $L_0$ regularization on the parameter θ with hypothesis h(.;θ) is

$$R(\theta; \lambda) = \frac{1}{N}\left(\sum_{i=1}^{N} L(h(x_i; \theta), y_i) + \lambda ||\theta||_0\right) \quad (1)$$

where λ is the Lagrange multiplier and L(.) is the loss function. L0 norms induce sparsest parameterizations because the penalty term associated with λ attempts to minimize the number of parameters entering into the model. Essentially, it penalizes all non-zero parameters the same cost, λ, regardless of the parameter value. This makes the problem combinatorial in nature and minimization with gradient descent hard due to its non-convex and discrete nature. We follow [1] approach to relax hard $L_0$ constraints with a smoother approximation. This is achieved in two steps 1- define an auxiliary variable, $z \, \epsilon \, (0,1)$, that acts as a mask on each parameter (this step is an exact transformation) and 2 – instead of learning auxiliary variable, z, we learn a probability distribution, q(s), over a variable, s, that acts as a continuous proxy for z and hence learnable through gradient descent type methods. The CDF of q(s), Q(s), is sampled using hard concrete distribution [2] with Gumbel-max trick for realization of mask variable, z. Fig. 3 shows a learned q(s) distribution over the parameters – parameters corresponding to $s \leq 0$ are likely to be masked in our derived model. Increasing λ shifts probability mass of q(s) to the left.

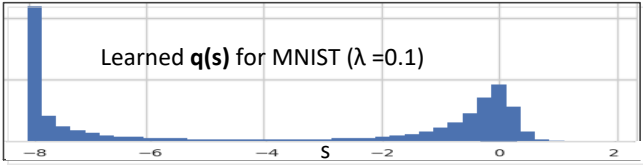

Fig. 3 Distribution of continuous variable s to proxy binary variable z

The relaxed-approximate empirical risk is Monte Carlo approximation to generally intractable expectation over hard samples over the probability distribution:

$$R(\tilde{\theta}, \phi; \lambda, L) = \frac{1}{L}\sum_{l=1}^{L}\left(\frac{1}{N}\sum_{i=1}^{N} L(h(x_i; \tilde{\theta}.z^{(l)}), y_i)\right) + \lambda \sum_{j=1}^{|\theta|}(1 - Q(s_j \leq 0|\phi_j)) \quad (2)$$

where L is the number of samples in Monte Carlo approximation, z(l) is the lth instance of mask associated with the parameters θ and $Q(s_j)$ is the cdf of the smooth probability distribution for parameter $s_j$. Here we have explicitly encoded regularized empirical risk, $R(\tilde{\theta}, \phi; \lambda, L)$, with λ and L for emphasis. We shall explore their impact on the design and performance of dense layer in neural network (NN) architecture.

## III. MNIST DATA

### A. Lagrange Multiplier

The value of Lagrange multiplier, λ, signifies the cost of including an additional parameter in the model. For typical regularization, its value is treated as a hyper-parameter that is tuned to optimize generalization/test error. Here, instead we shall explore its impact on joint value of generalization error and the number of parameters in our model (as proxy to HW complexity). We shall consider λ to range from 0 to ∞ - when λ=0, the model has maximum number of parameters and it corresponds to an unregularized/saturated model and when λ=∞, the number of parameters in the model is zero. Therefore, as we decrease its value, more parameters enter into our model as we trace a path in high dimensional parameter space.

Fig. 4 shows compression vs accuracy trade-off along the path for MNIST digit data.

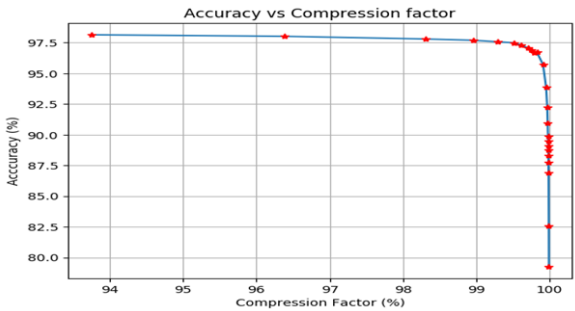

Fig. 4 Compression factor is % of parameters removed (masked) from the model. The curve shows a knee around 99% compression with accuracy loss of less than 1%

Fig. 5 is another rendition of same data inspired by rate distortion function in the information theory. The horizontal axis represents permissible distortion (100-%Accuracy) and the vertical axis represents ($\log_2$ of) the number of parameters (in bits) in our model. The figure shows a reduction in complexity by 4 bits (a factor of 16) for a cost of 0.1% accuracy. The curve allows us to trade accuracy with hardware complexity. Furthermore, across the breadth of the chart, we see a reduction in complexity by 13 bits (a factor of 8000) for 20% reduction in accuracy.

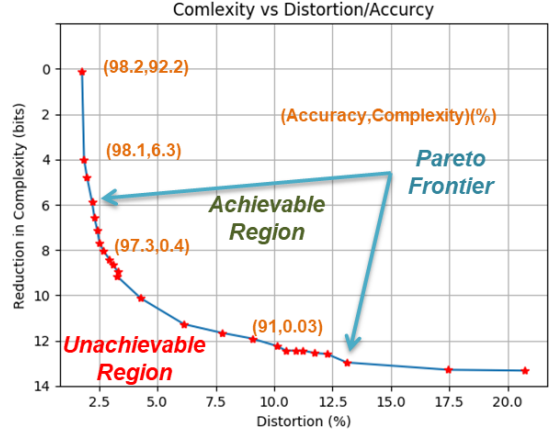

Fig. 5 Rate-Distortion function like curve – vertical axis represents hardware complexity in bits (a reduction by a single bit represents halving the model complexity). It shows a complexity reduction by a factor of 16 (4 bits) for accuracy loss of 0.1%

The curve represents Pareto frontier and is also known as isoquant in economics. The slope of the curve at any point measures the price of complexity in terms of accuracy. In other words, we are able to sort and score the parameters in our model relative to their importance to our desired objective (in our case – accuracy). Such sorting and scoring suggest models that respect given cost/complexity budget constraints.

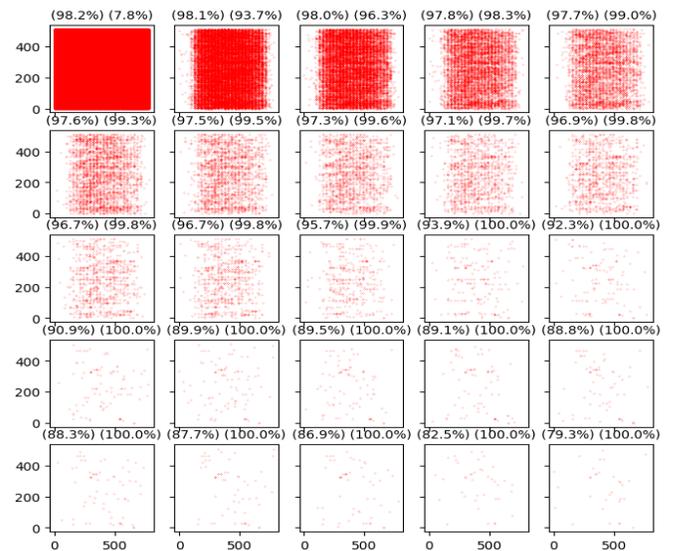

Fig. 6 Sparsity pattern induced by $L_0$ regularization for increasing values of λ

Fig. 6 shows the sparsity pattern induced by $L_0$ regularization for various values of λ. Horizontal axis of each subplot represents input node and the vertical axis represents an output node. A red dot indicates presence of an edge/weight between them. As we apply larger values of λ, we see that edges disappear. It is important to note the pattern of sparsity, first few subplots show absence edges on the margins and vertical gaps implying that the network has learnt to ignore pixel around the edges and achieved significant reduction at very low price in terms of accuracy.

Fig. 7 shows per class accuracy (vertical axis) for different values of λ. Unsurprisingly, all classed are not affected in the same way. For example, digit 5, 8 and 2 suffer significant loss of accuracy compare to digit 0 or 1. Hence, we can draw a rate distortion like curve for each digit and quantify impact of given parameter on per-class accuracy in the model suggesting accuracy equalization strategies.

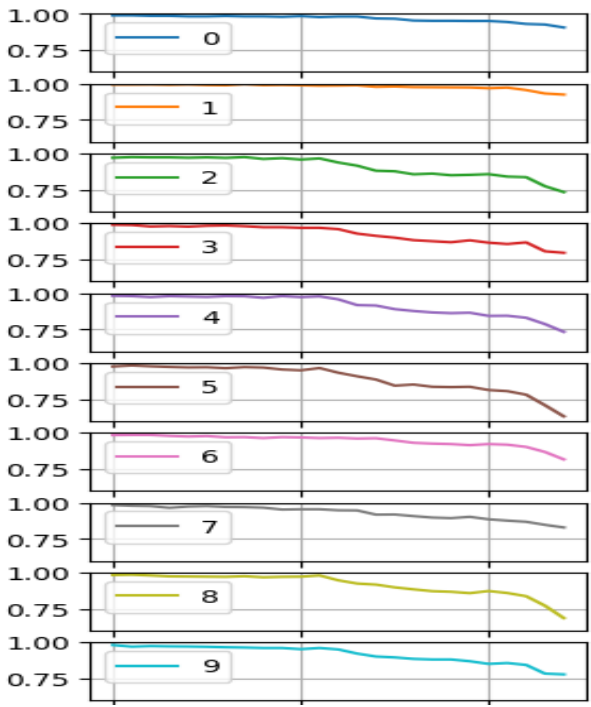

Fig. 7 Per-Class Accuracy against increasing values of Lagrange Multiplier λ

### B. Monte Carlo Approximation

As mentioned earlier, evaluation of our approximate loss function involves Monte Carlo approximation. A natural question may be to ask how many samples, L in (2), should be used? It turns out that samples are fairly correlated since only the parameters corresponding to s values near zeros are likely to change across sampling process. Fig. 8 and Fig. 9 show boxplot of the accuracy scores by aggregating various number sample model instances. We see that increasing the number of samples improves accuracy and it variance. We also see that a 2-instance model for λ=0.1 shows better accuracy than the baseline λ=0 model, suggesting improved performance while reducing complexity. For MNIST data, we don't see any significant improvement beyond 10 sample model. Prediction from various sample models can be combined to provide confidence level around each input data point as well.

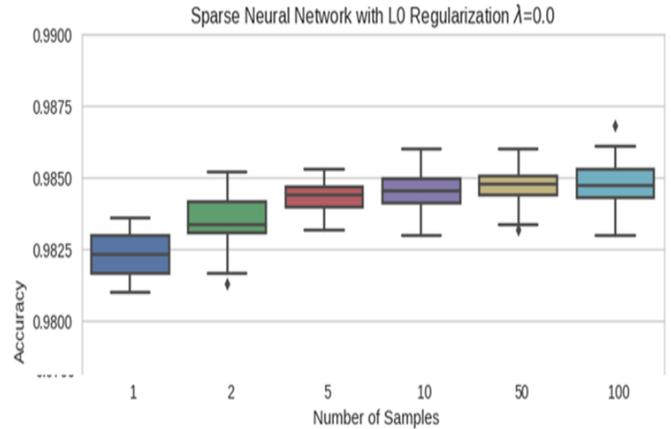

Fig. 8 Monte Carlo Approximation of Accuracy *λ=0*

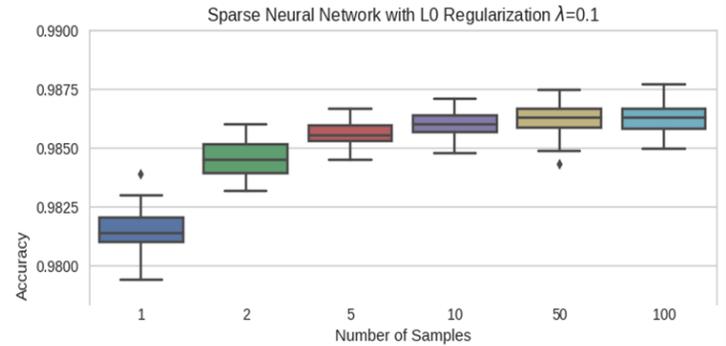

Fig. 9 Monte Carlo Approximation of Accuracy *λ=0.1*

In the above discussion, we aggregated accuracy scores from various sample models, but we can use more sophisticated ensembling techniques like boosting to further improve performance. We could also add complexity penalty in our Monte Carlo approximation for model selection.

### IV. MODREC DATA

Data was generated with GNU radio script "generate_RML2016.10a.py" located at /radioML/dataset on Github provided by [3]. The script generates 11 types of modulations at Signal to Noise Ratios (SNRs) ranging from -20dB to +18dB in steps of 2dB. For our training, we excluded signals with SNR less than -4dB. Each modulation and SNR combination contains 1000 waveforms of 128 complex samples. More details about dataset may be found in [3]. The network is composed of input, convolutional, convolutional, $L_0$ regularized fully connected, fully connected and softmax layers. The number of hidden units in all layers are part of hyper parameters to be tuned by validation data [4].

Fig. 10 shows the accuracy vs compress plot for modrec data. It is interesting to note initial improvement of accuracy while reducing the number of parameters. The generalization improvements are not surprising since we see similar improvements with dropouts. However, dropout mechanism does not lead to any complexity reduction. Note that we have specifically removed dropouts from the model given in [4] to isolate $L_0$ effects.

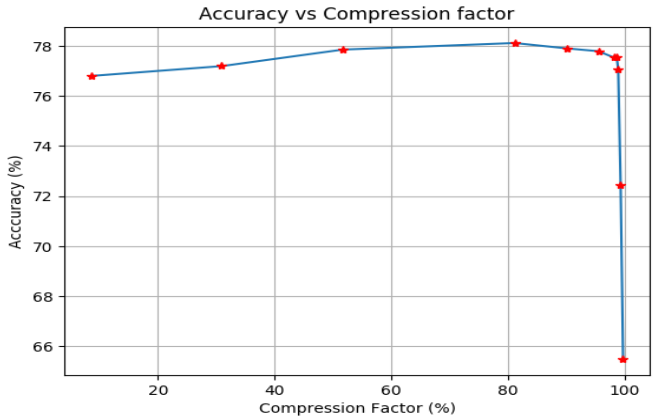

Fig. 10 Compression factor is % of parameters removed (masked) from the model. The curve shows a knee around 95% compression with accuracy ***gain*** of 1%

Fig. 11 shows the sparsity pattern for increasing values of λ for modrec data. The horizontal gaps imply subsampling of input complex signals with loss of significant performance.

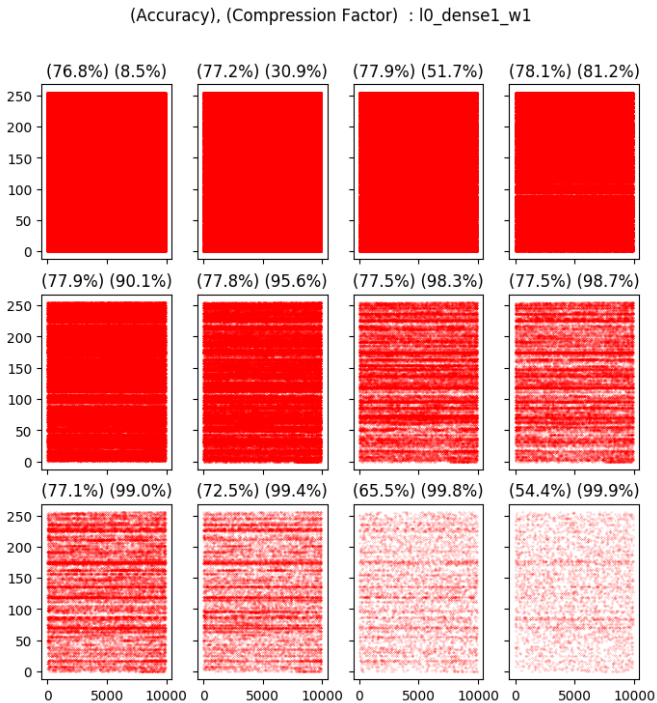

Fig. 11 Sparsity pattern induced by $L_0$ regularization for increasing values of λ

Fig. 12 and Fig. 13 show boxplot of accuracy scores for various size of aggregated ensemble models. The plot suggests improved accuracy score and variances for larger models. It also shows diminishing returns for larger ensembles. As mentioned earlier, other ensembling techniques like bagging and boosting may be employed to improve further design and performance.

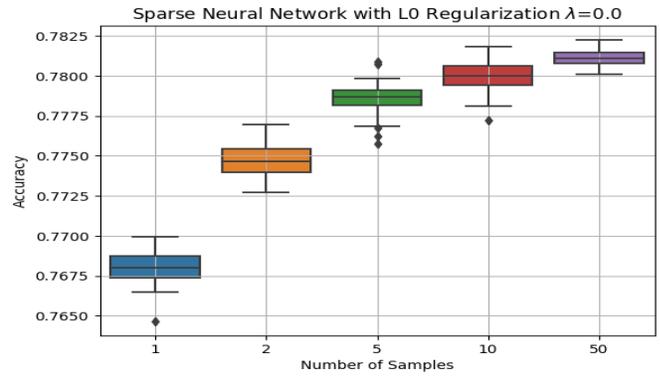

Fig. 12 Monte Carlo Approximation of Accuracy *λ=0.1*

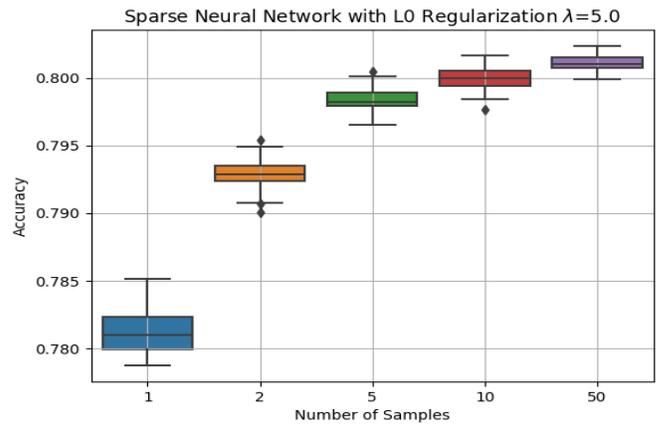

Fig. 13 Monte Carlo Approximation of Accuracy *λ=5*

## V. CONCLUSION

We used $L_0$ regularization scheme from [1] on MNIST and Modrec data for dense layer of the neural networks to trace out a path in parameter space with monotonically decreasing complexity as measured by L0 metric. We showed that when combined with ensembling techniques, the reduction in complexity may be achieved while improving performance. We validated generalization error by distilling our model to gradient boosted trees.